\documentclass[conference]{IEEEtran}
\IEEEoverridecommandlockouts
\usepackage{cite}
\usepackage{amsmath,amssymb,amsfonts}
\usepackage{algorithmic}
\usepackage{graphicx}
\usepackage{booktabs}
\usepackage{textcomp}
\usepackage{xcolor}
\usepackage{multirow}
\usepackage{mathtools}
\usepackage{graphbox}
\usepackage{float}

\usepackage{tikz}
\usepackage{textcomp}
\usepackage{hyperref}
\usepackage{lipsum}

\newcommand\copyrighttext{%
  \footnotesize \textcopyright 2023 IEEE. Personal use of this material is permitted.
  Permission from IEEE must be obtained for all other uses, in any current or future
  media, including reprinting/republishing this material for advertising or promotional
  purposes, creating new collective works, for resale or redistribution to servers or
  lists, or reuse of any copyrighted component of this work in other works.
  DOI: \href{https://doi.org/10.1109/ICVR57957.2023.10169359}{10.1109/ICVR57957.2023.10169359}}
\newcommand\copyrightnotice{%
\begin{tikzpicture}[remember picture,overlay]
\node[anchor=south,yshift=10pt] at (current page.south) {\fbox{\parbox{\dimexpr\textwidth-\fboxsep-\fboxrule\relax}{\copyrighttext}}};
\end{tikzpicture}%
}

\def\BibTeX{{\rm B\kern-.05em{\sc i\kern-.025em b}\kern-.08em
    T\kern-.1667em\lower.7ex\hbox{E}\kern-.125emX}}

\begin{document}

\title{Dense Voxel 3D Reconstruction Using a Monocular Event Camera}

\author{\IEEEauthorblockN{Haodong Chen}
\IEEEauthorblockA{\textit{School of Computer Science}\\
\textit{The University of Sydney} \\
Sydney, Australia \\
haodong.chen@sydney.edu.au}
\and
\IEEEauthorblockN{Vera Chung}
\IEEEauthorblockA{\textit{School of Computer Science}\\
\textit{The University of Sydney} \\
Sydney, Australia \\
vera.chung@sydney.edu.au}
\and
\IEEEauthorblockN{Li Tan}
\IEEEauthorblockA{\textit{School of Computer Science and Engineering} \\
\textit{Beijing Technology and Business University}\\
Beijing, China \\
tanli@btbu.edu.cn}
\and
\IEEEauthorblockN{Xiaoming Chen}
\IEEEauthorblockA{\textit{School of Computer Science and Engineering} \\
\textit{Beijing Technology and Business University}\\
Beijing, China \\
xiaoming.chen@btbu.edu.cn \\ (Corresponding Author)}
}

\maketitle
\copyrightnotice

\begin{abstract}
Event cameras are sensors inspired by biological systems that specialize in capturing changes in brightness. These emerging cameras offer many advantages over conventional frame-based cameras, including high dynamic range, high frame rates, and extremely low power consumption. Due to these advantages, event cameras have increasingly been adapted in various fields, such as frame interpolation, semantic segmentation, odometry, and SLAM. However, their application in 3D reconstruction for VR applications is underexplored. Previous methods in this field mainly focused on 3D reconstruction through depth map estimation. Methods that produce dense 3D reconstruction generally require multiple cameras, while methods that utilize a single event camera can only produce a semi-dense result. Other single-camera methods that can produce dense 3D reconstruction rely on creating a pipeline that either incorporates the aforementioned methods or other existing Structure from Motion (SfM) or Multi-view Stereo (MVS) methods. In this paper, we propose a novel approach for solving dense 3D reconstruction using only a single event camera. To the best of our knowledge, our work is the first attempt in this regard. Our preliminary results demonstrate that the proposed method can produce visually distinguishable dense 3D reconstructions directly without requiring pipelines like those used by existing methods. Additionally, we have created a synthetic dataset with $39,739$ object scans using an event camera simulator. This dataset will help accelerate other relevant research in this field.
\end{abstract}


\begin{figure*}[!ht]
\centering
\includegraphics[width=\linewidth]{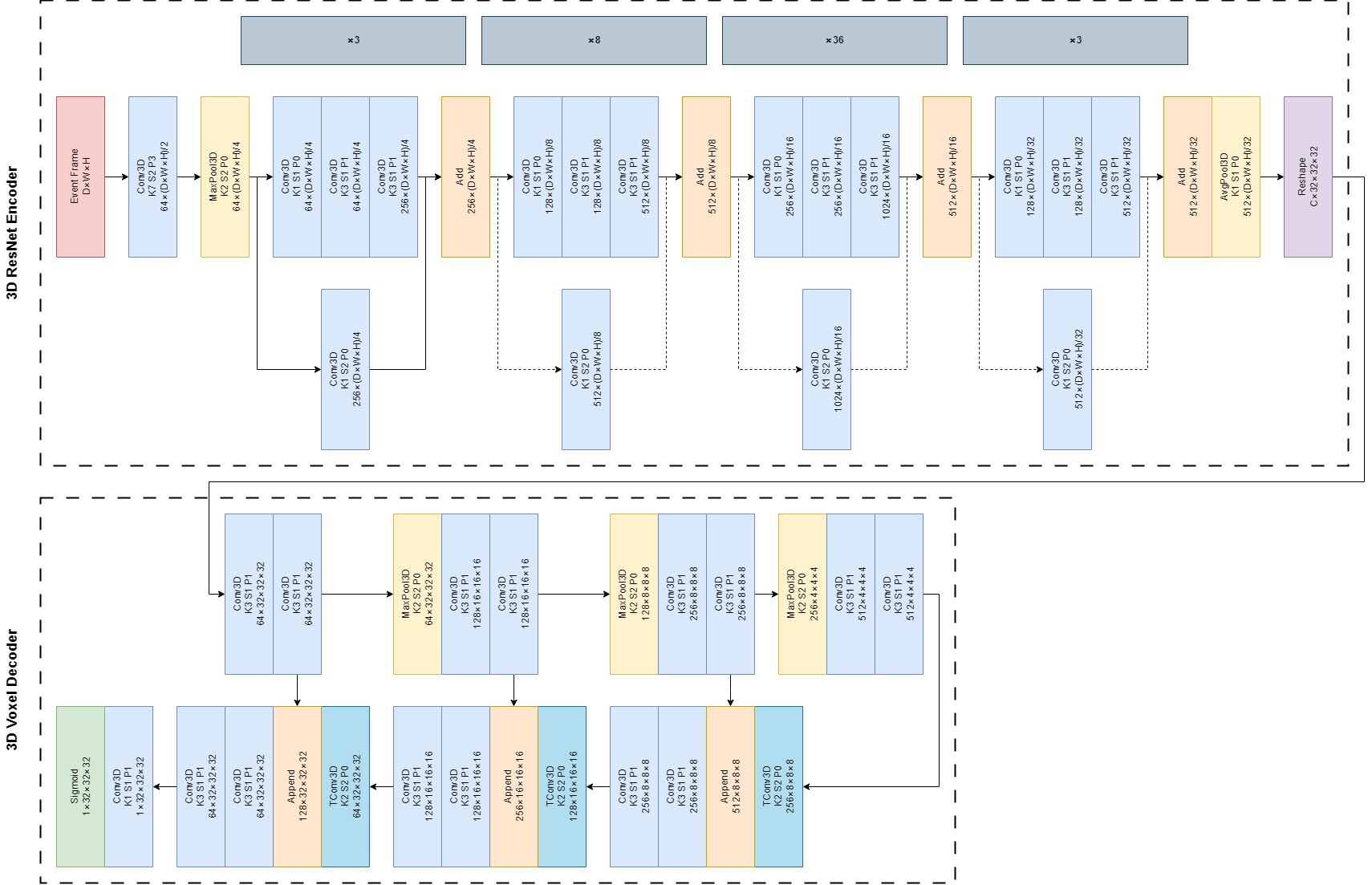}
\caption{The architecture of the proposed E2V. The encoder is a modified version of ResNet-152 with all 2D convolutions replaced by 3D convolutions. }
\label{fig_model_structure}
\end{figure*}

\section{Introduction}
Event cameras are a new type of imaging sensor inspired by the biological retina \cite{lichtsteiner2008128}. Unlike conventional cameras, they capture only changes in brightness asynchronously in each pixel of the sensor array. This creates a data stream that includes timestamps, pixel coordinates, and polarities. Compared to conventional cameras, event cameras have several advantages, including high dynamic range, high frame rates, and extremely low power consumption. Therefore, event cameras have increasingly been used in many fields, such as frame interpolation, object detection, semantic segmentation, odometry, and SLAM, with varying degrees of success.

Event cameras have also been applied in the field of 3D reconstruction. However, current top-performing approaches require multiple event cameras to perform depth estimation \cite{kogler2011address}, \cite{ieng2018neuromorphic}, and then create a 3D reconstruction from the estimated depth maps. Although it is possible to use one event camera for the same task, these methods can only create semi-dense reconstructions \cite{rebecq2018emvs}, \cite{kim2016real}. Other monocular methods that can produce dense reconstruction results all require pipelines \cite{DBLP:journals/corr/abs-2012-05214}, \cite{xiao2022event} that incorporate multiple existing methods. In this paper, we attempt to address this problem by proposing a method that generates a dense 3D voxel grid using simulated event camera data. When compared to other existing methods in this field, our proposed method only requires a single event camera without requiring any intrinsic or extrinsic parameters of the camera, and it can directly output a dense 3D voxel reconstruction. 

Our preliminary results indicate that our proposed method can create visually recognizable 3D voxel reconstructions. Based on our tests, the reconstructed voxel results from our proposed model are comparable to some of the existing RGB image-based 3D reconstruction methods after training on a small dataset of $832$ data for $100$ epochs. In future work, we aim to achieve better results than traditional RGB-based algorithms by training the proposed method with a larger dataset. 

Our main contributions are:
\begin{itemize}
    \item We propose the first dense 3D reconstruction method using only one event camera without relying on pipelines.
    \item The proposed method is capable of generating dense voxel 3D reconstruction, and the results can be visually comparable to traditional image-based methods. 
    \item We constructed the first large-scale dataset for event-based 3D reconstruction with $39,739$ object scans. 
\end{itemize}

The remaining sections of this paper are structured as follows. We review related work in Section \ref{sec_related_works}, explain the current state of event-based 3D reconstruction, and mention several image-based methods used for comparison in the experiment. In Section \ref{sec_methodology}, we explain the event representations used in this paper, introduce the detailed architecture of our proposed method and give details on the setup, camera configuration, and workflow of the dataset creation process. In Section \ref{sec_experiments}, we provide detailed network configuration and hyper-parameter settings and then evaluate the experiment result. The full result is shown in Table \ref{results_table}, and some reconstruction examples can be seen in Fig. \ref{E2V_visual_results}. In the end, we conclude the work and discuss possible improvements to the proposed method for our future work.

\begin{table*}[!ht]
\centering
\renewcommand{\arraystretch}{1.1}
\caption{Experimental Results of IoU for the proposed E2V}
\label{results_table}
\resizebox{0.9\textwidth}{!}{%
\begin{tabular}{@{}llllllllllllllll@{}}
\toprule
 & Data Size & airplane & bench & cabinet & car & chair & display & lamp & speaker & rifle & sofa & table & telephone & watercraft & Overall \\ \midrule
Training Data & 832 & 0.652 & 0.697 & 0.879 & 0.710 & 0.740 & 0.817 & 0.681 & 0.807 & 0.597 & 0.746 & 0.801 & 0.882 & 0.643 & 0.742 \\
Testing Data & 104 & 0.485 & 0.293 & 0.405 & 0.520 & 0.261 & 0.310 & 0.158 & 0.190 & 0.512 & 0.400 & 0.297 & 0.377 & 0.372 & 0.352 \\
Large Testing IoU & 8770 & 0.424 & 0.364 & 0.344 & 0.518 & 0.234 & 0.310 & 0.182 & 0.240 & 0.426 & 0.355 & 0.250 & 0.488 & 0.361 & 0.346 \\ \bottomrule
\end{tabular}%
}
\end{table*}

\begin{table*}[!ht]
\centering
\renewcommand{\arraystretch}{1.1}
\caption{Experimental Results for the proposed E2V with F-Score@$20\%$}
\label{results_table_fscore}
\resizebox{0.9\textwidth}{!}{%
\begin{tabular}{@{}llllllllllllllll@{}}
\toprule
 & Data Size & airplane & bench & cabinet & car & chair & display & lamp & speaker & rifle & sofa & table & telephone & watercraft & Overall \\ \midrule
Training Data & 832 & 0.291 & 0.268 & 0.167 & 0.191 & 0.168 & 0.264 & 0.283 & 0.160 & 0.375 & 0.185 & 0.220 & 0.379 & 0.257 & 0.247 \\
Testing Data & 104 & 0.182 & 0.091 & 0.068 & 0.117 & 0.050 & 0.079 & 0.082 & 0.040 & 0.310 & 0.105 & 0.081 & 0.147 & 0.150 & 0.116 \\
Large Testing Data & 8770 & 0.164 & 0.118 & 0.068 & 0.116 & 0.054 & 0.078 & 0.087 & 0.052 & 0.252 & 0.078 & 0.067 & 0.205 & 0.112 & 0.112 \\ \bottomrule
\end{tabular}%
}
\end{table*}

\section{Related Works}\label{sec_related_works}
In this section, we review the current event-based 3D reconstruction methods and talk about their strengths and weaknesses. We also review a few popular image-based 3D reconstruction methods that were used to compare with our proposed method. 

\subsection{Depth Estimation Based Event 3D Reconstruction}
The current mainstream approach toward event 3D reconstruction is based on depth estimation. Methods following this approach can produce good 3D reconstruction results in a short amount of time with low computation requirements in most cases. This approach can be divided into two categories based on the number of cameras used, either singular or multiple (usually two) cameras. A review on event-based vision has also covered this topic in great detail \cite{gallego2020event}. 

A typical stereo setup consists of two rigidly attached event cameras with synchronized clocks. Most methods under this category can achieve instantaneous depth map estimation. Some stereo methods convert raw events into event frames \cite{kogler2011address} or time surfaces \cite{ieng2018neuromorphic} and following classical image-based stereo methods \cite{hartley2003multiple}, are able to solve the epipolar matching problem and triangulate the location of the 3D points. Other stereo methods are able to exploit the simultaneity and temporal correlations of the events to achieve the same goal \cite{ieng2018neuromorphic}, \cite{kogler2011event}. By adding additional regularity constraints \cite{piatkowska2013asynchronous}, using brute-force space-sweeping method on dedicated GPU \cite{zhu2018realtime} or pairing event cameras with neuromorphic processors \cite{dikov2017spiking}, better depth estimation can also be achieved. 

Monocular depth estimation, on the other hand, only consists of a single event camera. Because this approach can not exploit the temporal correlation between events across multiple sensors, the methods can only estimate the depth by moving the camera over time \cite{rebecq2018emvs}, \cite{kim2016real} and are only capable of producing semi-dense 3D reconstruction. Though it is possible to obtain dense 3D reconstruction using a single camera, it often requires additional data or needs multiple post-processing methods in a pipeline-like formation \cite{DBLP:journals/corr/abs-2012-05214}, \cite{xiao2022event}. 

What both depth estimation based 3d reconstruction methods have in common is that they both require additional post-processing steps such as marching cubes or Delaunay triangulation in order to obtain 3D reconstructed representations (3D mesh, point cloud, or voxel, etc.). On the other hand, our method only requires a single camera and can directly produce a dense voxel reconstruction. 

\subsection{Conventional Image Based 3D Reconstruction}
The dataset we created is based on ShapeNet \cite{shapenet2015}, a widely-used large-scale 3D model repository for RGB image-based 3D reconstruction. Since most existing event-based 3D reconstruction methods use multiple cameras and depth estimation to solve the reconstruction problem, comparing our method with theirs is difficult. Therefore, we selected a few image-based 3D reconstruction methods as references. The top-performing methods in multi-view 3D object reconstruction currently are 3D-R2N2 \cite{choy20163d}, AttSets \cite{yang2020robust} Pix2Vox++ \cite{xie2020pix2vox++} and EVolT \cite{wang2021multi}, with EVolT as the current state-of-the-art on ShapeNet \cite{shapenet2015}.

\section{Methodology}\label{sec_methodology}
\subsection{Event Representation}
Since event data differs significantly from conventional image-based data, we need to apply preprocessing methods to transform the event data into a format that is compatible with existing deep learning models. Typical event data recorded on an $M \times N$ sensor for a duration of $T$ can be represented as follows:

\begin{equation}
e = \{\varepsilon_i\}^I_{i=1}
\end{equation}
\begin{equation}
\varepsilon_i = (\mathbf{x}_i, t_i, p_i)
\end{equation}
\begin{equation}
\mathbf{x}_i = (x_i, y_i)  
\end{equation}

where $e$ is the entire event sequence with $I \in \mathbb{N}$ events. $\varepsilon_i$ represents an individual event and $\mathbf{x}_i$ is the coordinate on the sensor with $x_i \in [0,\cdots,M]$ and $y_i \in [0,\cdots,N]$. $t_i \in \mathbb{R}^+$ is the time stamp with $t_i \leq t_{i+1}$ for $i < I$ and $t_I = T$. $p_i \in \{1, -1\}$ is the polarity that denotes the brightness change. 

In this paper, we used event frames to represent events so that they can be fed into deep learning models. The $k^{th}$ event frame can be obtained by accumulating events onto a 2D plane $F_k(u, v) \in \{0,1\}^{M \times N}$ within a set time interval $\Delta_t$.

\begin{equation}
    F = \{F_k(u, v)\}
    \begin{cases}
        1 \text{ if } (u,v) \in f_k \\
        0 \text{ otherwise}
    \end{cases}
\end{equation}

\begin{equation}
f = \{f_k\}
    \begin{cases}
        \begin{rcases}
            f_k \cup \{ \mathbf{x}_i \} \\
            i=i+1
        \end{rcases}
        \text{if } t_i - t_j \leq \Delta_t \\
        \begin{rcases}
        k = k + 1 \\
        f_k = \emptyset \\
        j = i \\
        \end{rcases}
        \text{ otherwise}
    \end{cases}
    \begin{cases}
        k = 0\\
        f_k = \emptyset \\
        i = 1 \\
        j = i
    \end{cases}
\end{equation}

\subsection{Framework}
In this paper, we propose a new deep learning model called Event to Voxel (E2V), consisting of a 3D event frame encoder and a 3D voxel decoder. Initially, we could not use conventional CNN models as the encoder due to the large size of our input data ($batch\_size \times 100 \times 256 \times 256$). The model size would quickly increase with each additional convolutional layer. Therefore, we chose ResNet as the encoder because it uses a convolutional layer with a large kernel size at the beginning, which greatly reduces the size of the input data. The encoder is a modified version of ResNet-152 \cite{he2016deep}, where all 2D convolutional layers have been replaced with 3D convolutions. The encoder takes preprocessed event frames with a shape of $D\times W \times H$ and outputs them to a hidden layer with a shape of $C \times 32 \times 32 \times 32$. Here $D$ is the number of frames, $W$ and $H$ are the width and height of the frame, and $C$ is the hidden channel size calculated based on the input shape. 

The 3D voxel decoder takes the encoded hidden layer and outputs the 3D voxel representation of the captured object. The UNet-like \cite{ronneberger2015u} decoder is inspired by the refiner in Pix2Vox++ \cite{xie2020pix2vox++}, which was used to improve the decoder's output. However, in our case, the decoder more closely resembles the original UNet architecture. By going through examples of preprocessed event data and 3D voxel labels, the model learns the correlation between them, allowing it to map out the voxel representation of newly captured event data of the target object. 

\subsection{Dataset}
According to a survey conducted by G. Gallego et al. in 2020 \cite{gallego2020event}, the field of event-based 3D reconstruction currently lacks a comprehensive dataset. Furthermore, because our method diverges from conventional depth estimation-based approaches, comparing it with existing methods becomes even more challenging. To address this, we constructed a synthetic dataset using ShapeNet \cite{shapenet2015}, a commonly used 3D dataset for traditional RGB image-based 3D reconstruction that contains over $50,000$ 3D models.

\begin{figure}[!ht]
\centering
\includegraphics[width=0.8\linewidth]{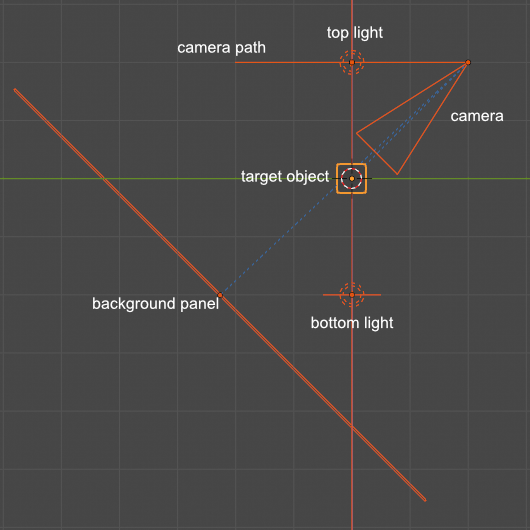}
\caption{Side view of the camera setup.}
\label{fig_camera_setup}
\end{figure}

\begin{figure}[!ht]
\centering
\includegraphics[width=0.8\linewidth]{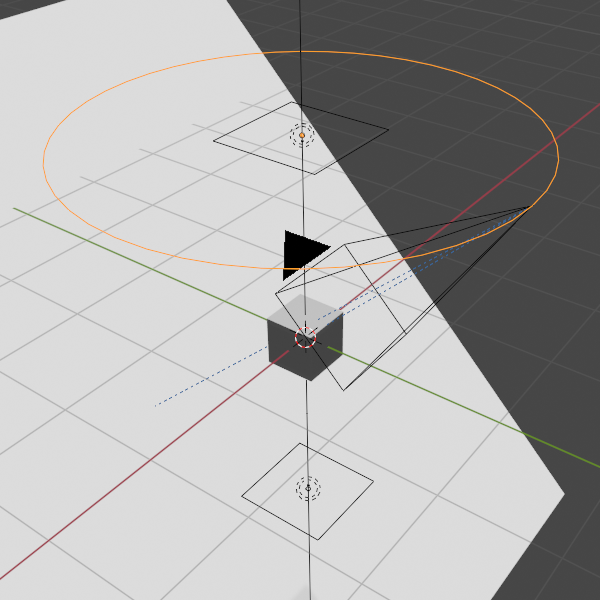}
\caption{Perspective view of the camera setup}
\label{fig_camera_setup_sideview}
\end{figure}
Our synthetic dataset is created using a two-step process. First, we capture high frame rate video using 3D modeling software like Blender. Within the 3D scene, we define a circular camera path with a starting radius of $4m$. The camera moves along this path while always locked onto the object. The camera path's position changes from $+2m$ to $-2m$ on the z-axis over time. The camera path's diameter gradually increases from $4m$ to $6m$ while moving towards $0m$ on the z-axis, then scales back to $4m$ while moving from $0m$ to $-2m$. This ensures that the recorded video can capture all sides of the target object while keeping the distance from the camera to the object near constant.

We used three light sources to illuminate the scene, including sunlight with $2W/m^2$ strength and two $10 W$ $1m \times 1m$ square lights placed directly on top and below the target object with a $2m$ distance. Additionally, we use a white background panel that moves with the camera on the opposite side to ensure that the background color of the video remains constant during recording. All videos were recorded at $240$fps within $0.5$ seconds using a virtual camera with $80 mm$ focal length and spatial resolution of $512 \times 512$ pixels.

After capturing the video data, we feed the videos into the Video to Events model\cite{gehrig2020video}, which provides us with the simulated event data. All 3D models from ShapeNet are normalized to fit within a $1m^3$ volume. For comparison purposes, we converted the 3D meshes into $32^3$ voxels and used them as labels following the vocalization process from Pix2Vox++ \cite{xie2020pix2vox++}. 

In the end, we generate $39,739$ event data scan from $13$ object categories. Due to limited computation power, for this experiment, we randomly selected $832$ training data, $104$ validation data, and $104$ testing data to verify the viability of our proposed method for this experiment. 

\begin{figure*}[!ht]
\renewcommand{\arraystretch}{1.1}
\setlength\tabcolsep{12pt}
\centering
\begin{tabular}{cccccc}
 & airplane & bench & cabinet & car & chair \\
Ground Truth & 
\includegraphics[align=c, width=0.12\linewidth]{
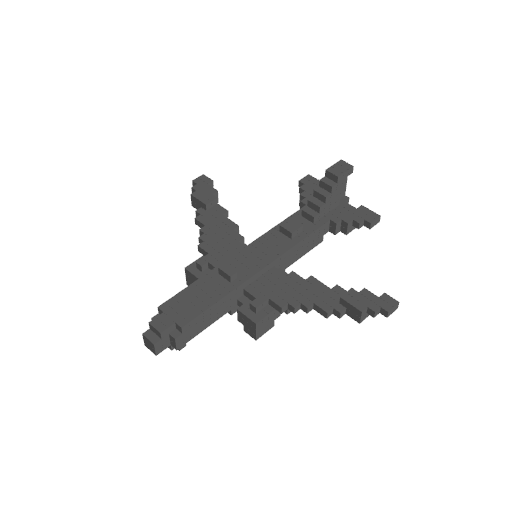} &
\includegraphics[align=c, width=0.12\linewidth]{
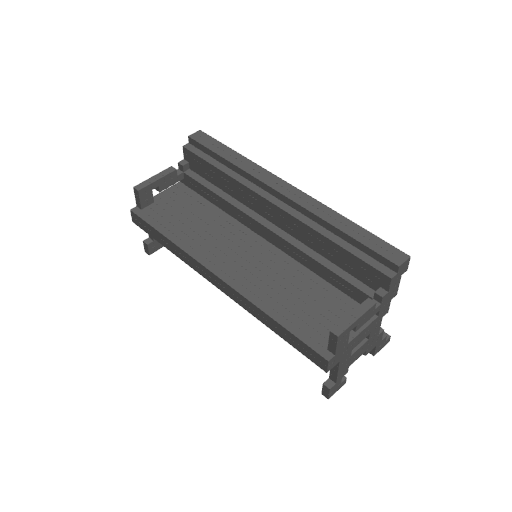} &
\includegraphics[align=c, width=0.12\linewidth]{
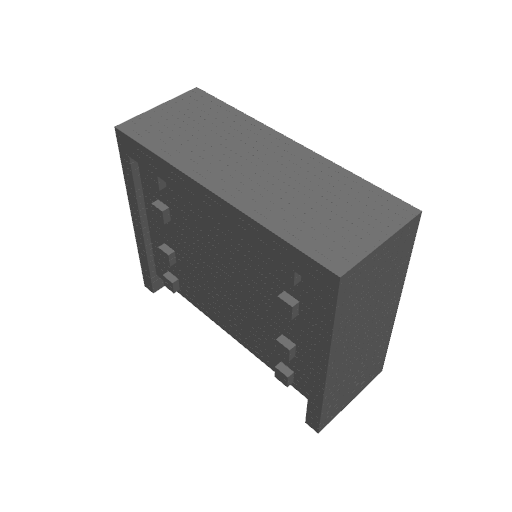} &
\includegraphics[align=c, width=0.12\linewidth]{
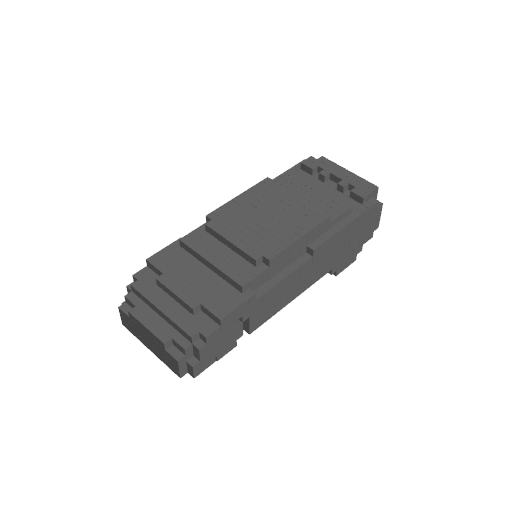} &
\includegraphics[align=c, width=0.12\linewidth]{
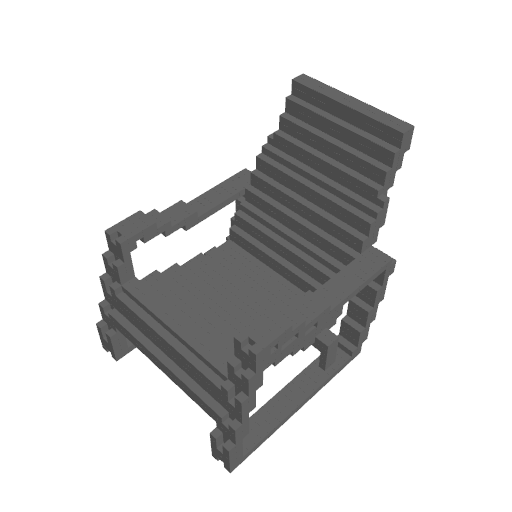} \\
E2V Result &
\includegraphics[align=c, width=0.12\linewidth]{
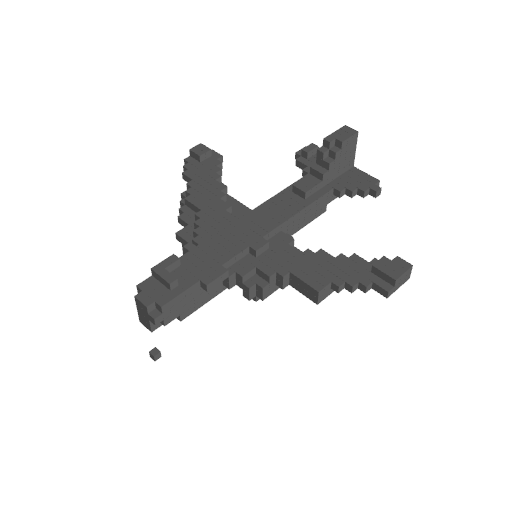} &
\includegraphics[align=c, width=0.12\linewidth]{
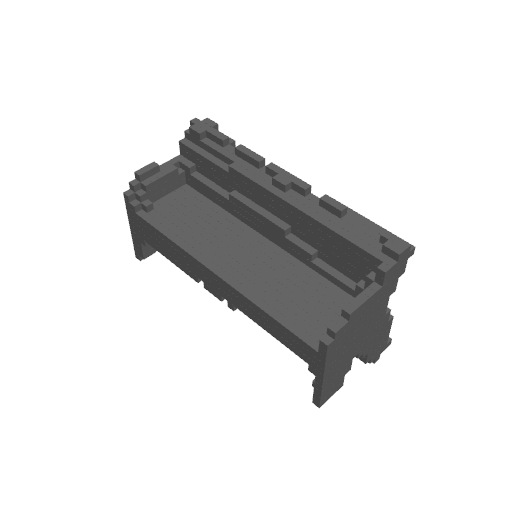} &
\includegraphics[align=c, width=0.12\linewidth]{
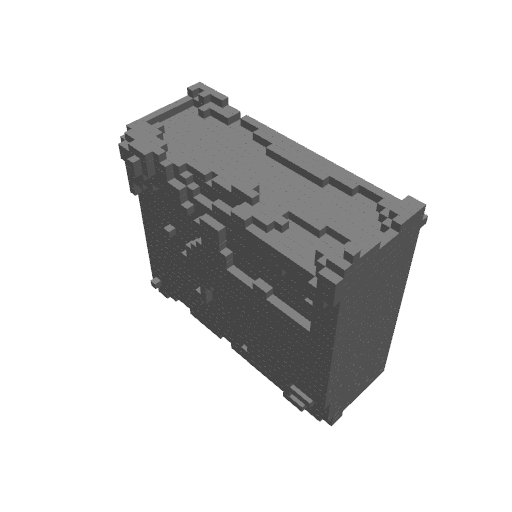} &
\includegraphics[align=c, width=0.12\linewidth]{
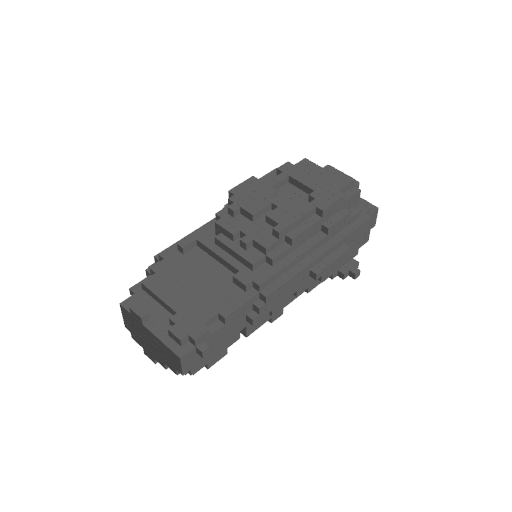} &
\includegraphics[align=c, width=0.12\linewidth]{
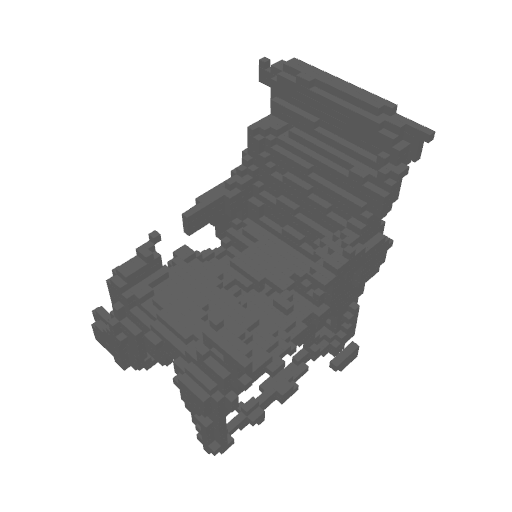} \\

 & display & lamp & speaker & sofa & table \\
Ground Truth & 
\includegraphics[align=c, width=0.12\linewidth]{
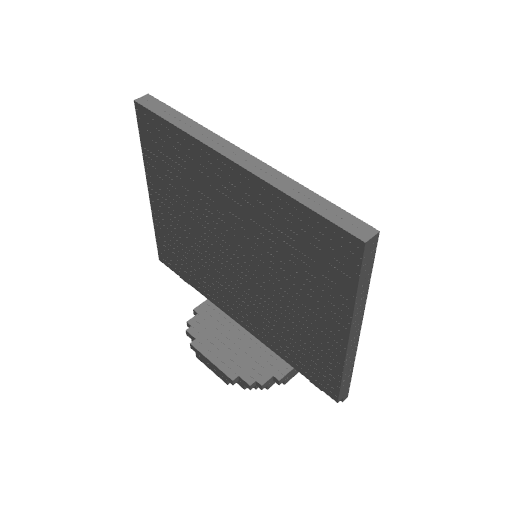} &
\includegraphics[align=c, width=0.12\linewidth]{
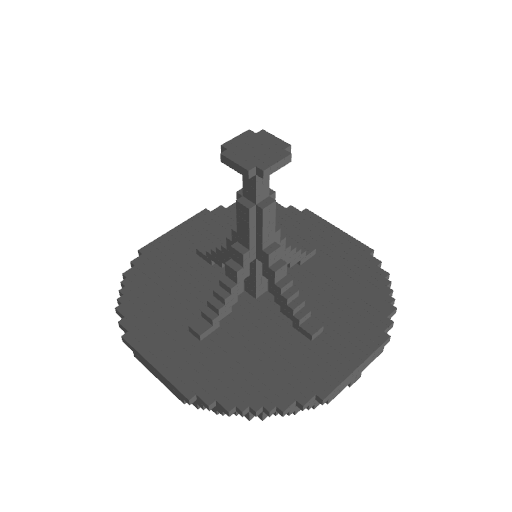} &
\includegraphics[align=c, width=0.12\linewidth]{
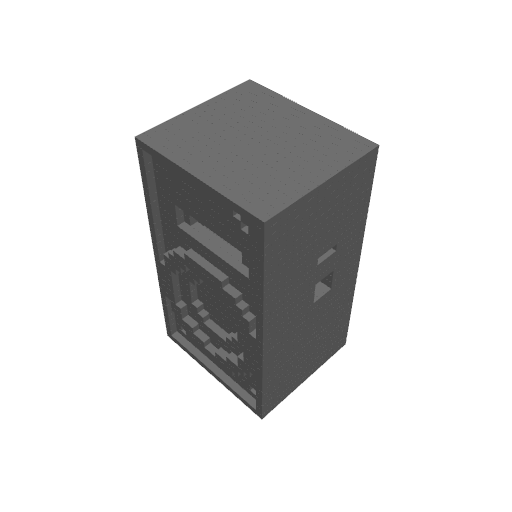} &
\includegraphics[align=c, width=0.12\linewidth]{
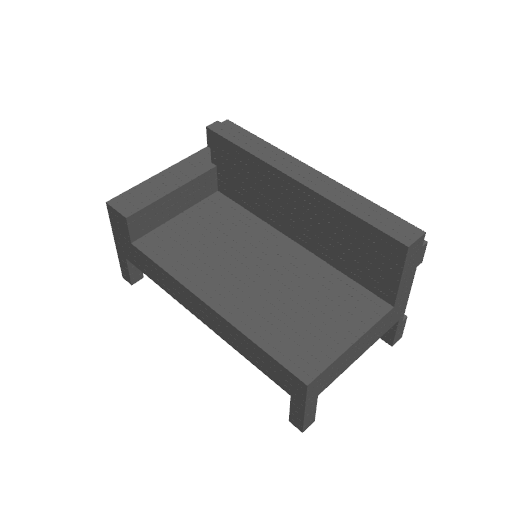} &
\includegraphics[align=c, width=0.12\linewidth]{
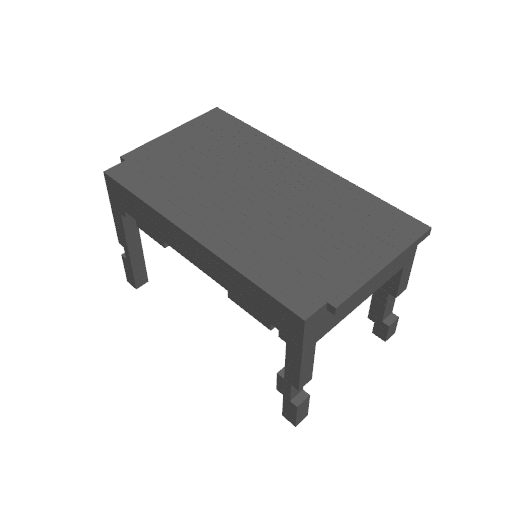} \\
E2V Result &
\includegraphics[align=c, width=0.12\linewidth]{
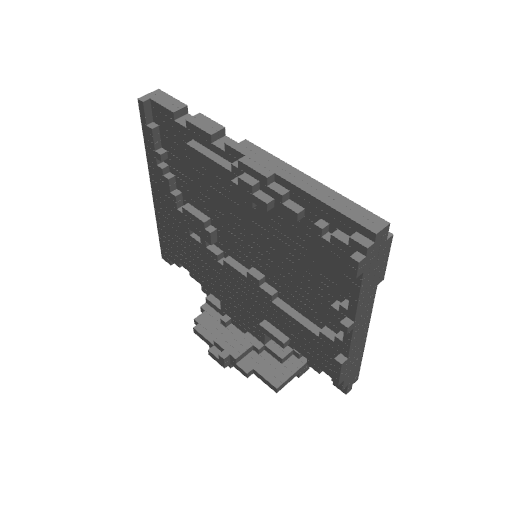} &
\includegraphics[align=c, width=0.12\linewidth]{
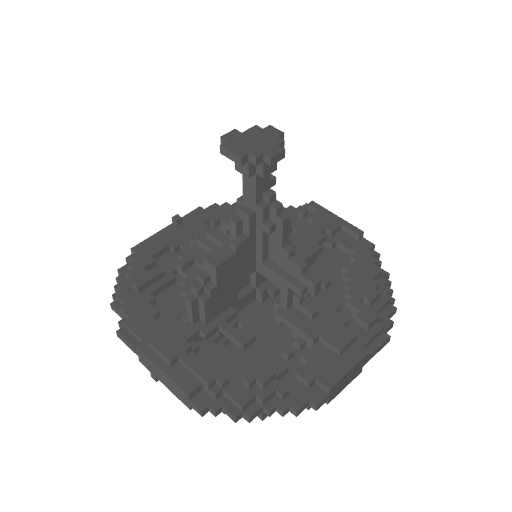} &
\includegraphics[align=c, width=0.12\linewidth]{
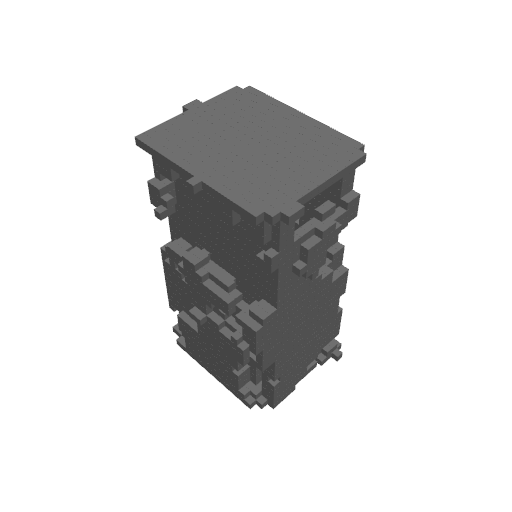} &
\includegraphics[align=c, width=0.12\linewidth]{
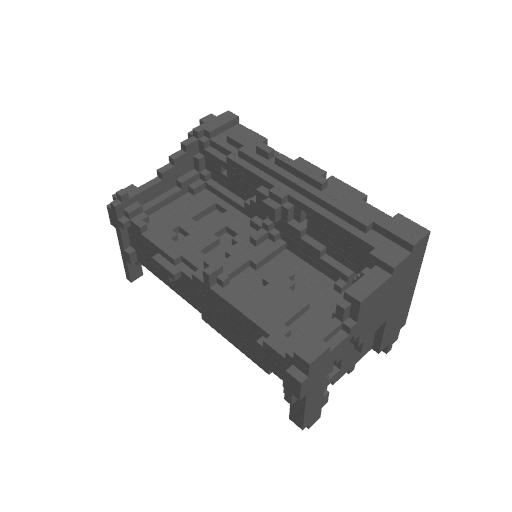} &
\includegraphics[align=c, width=0.12\linewidth]{
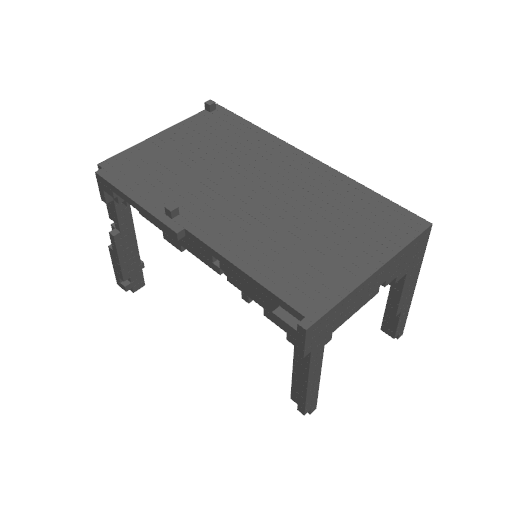} \\

\end{tabular}
\caption{3D reconstruction results for the proposed E2V model on test data.}
\label{E2V_visual_results}
\end{figure*}

\section{Experiments}\label{sec_experiments}
For this experiment, the proposed E2V model was trained for $100$ epochs with a batch size of $5$, using AdamW optimizer \cite{loshchilov2017decoupled} with $0.00001$ learning rate and mean value binary cross entropy loss function. The loss function can be formally defined as:

\begin{equation}
CE = -\frac{1}{N} \sum^N_{n=1}
[v_n \cdot log(p_n) + (1 - v_n) \cdot log(1 - p_n) ]
\end{equation}

where $N$ is the total number of voxels, $v_n$ is $n_{th}$ ground truth voxel point and $p_n$ is corresponding predicted voxel value.

To preprocess event data into event frames, we used $\Delta_t=0.005$. Since all event data were recorded within a $0.5s$ duration, every event sequence could generate $100$ event frames. We then scaled the event frame size from $512^3$ to $256^3$ to reduce computation costs. In the end, the input shape for E2V becomes $batch\_size \times 100 \times 256 \times 256$. The final model used in this experiment has $149,155,905$ parameters. 

\subsection{Evaluation}
To evaluate the reconstruction result, we selected two commonly used metrics in this field \cite{xie2020pix2vox++}, \cite{wang2021multi}: Intersection-over-Union (IoU) and F-Score. IoU can be calculated as follows:

\begin{equation}
    \text{IoU} = \frac{\sum_{i,j,k} I\left(p\left(i,j,k\right) >t\right) \cdot I\left(v\left(i,j,k\right)\right)}
               {\sum_{i,j,k} I\left(I\left(p\left(i,j,k\right) >t\right) + I\left(v\left(i,j,k\right)\right)\right)}
\end{equation}

where $p(i,j,k)$ is the predicted probability, $v(i,j,k)$ is the ground truth, $I(\cdot)$ is the indicator function and $t$ is the voxelization threshold. 

F-Score, on the other hand, can be interpreted as the distance between object surfaces \cite{wang2021multi}. It can be calculated as follows: 

\begin{equation}
\text{F-Score}(d) = \frac{2P(d)R(d)}{P(d)+R(d)}
\end{equation}
\begin{equation}
P(d) = \frac{1}{|\mathcal{R}|} \sum_{r\in \mathcal{R}} 
\left[ \min_{g\in \mathcal{G}} \left\| g-r \right\| < d \right]
\end{equation}
\begin{equation}
R(d) = \frac{1}{|\mathcal{G}|} \sum_{g\in \mathcal{G}} 
\left[ \min_{r\in \mathcal{R}} \left\| g-r \right\| < d \right]
\end{equation}

where $[\cdot]$ is the Iverson bracket. $\mathcal{G}$ and $\mathcal{R}$ are the ground truth and reconstructed point sets respectively. 

After the training process, the model was able to produce visually distinguishable 3D reconstruction results which close to the ground truth, as shown in Figure \ref{E2V_visual_results}. This suggests that the model is capable of inferring voxels on new event data. 

\begin{table}[!ht]
\centering
\renewcommand{\arraystretch}{1.1}
\caption{mIoU scores for RGB-based models}
\label{results_table_rgb}
\resizebox{0.7\linewidth}{!}{
\begin{tabular}{@{}lcl@{}}
\toprule
 & \multicolumn{1}{l}{Training Data Size} & mIoU \\ \midrule
3D-R2N2 & \multirow{4}{*}{$\sim$30,000} & 0.636 \\
AttSets &  & 0.693 \\
Pix2Vox++ &  & 0.706 \\
EVolT &  & 0.735 \\
E2V (Ours) & 832 & 0.346 \\ \bottomrule
\end{tabular}%
}
\end{table}

Table \ref{results_table} and \ref{results_table_fscore} present detailed experimental results of our proposed E2V model for event-based data. Table \ref{results_table_rgb}  shows the mIoU scores from other top-performing RGB-based methods. When comparing our event-based model to RGB-based models, a noticeable gap is observed in their IoU performance.  This disparity is expected, as RGB images contain more information, and the models were trained on a larger dataset. The gap between event and RGB-based methods is even more significant in terms of F-Score. RGB-based methods usually achieve an F-Score between $0.4$ and $0.5$ with a threshold of $1\%$. However, if we apply the same threshold to our proposed method, the F-Score will drop to zero. Therefore, a direct comparison of these two approaches is not feasible. Instead, we set the threshold at $20\%$ as a benchmark for future event-based 3D reconstruction methods.

\section{Conclusion and Future Works}
In this paper, we propose a novel dense 3D voxel reconstruction model that uses monocular event-based data for the first time. Our results show that event-based 3D reconstruction is a promising research direction for future VR applications. Moreover, our event dataset, which has over $39,739$ object scans, will be publicly available to help accelerate research in this field.

However, we are aware of some limitations in our work. Due to the lack of computational resources, we were only able to test our method on a small dataset. While the produced results cannot fully match the current state-of-the-art image-based methods, we believe that the performance of event-based 3D reconstruction can be significantly improved in the next version. We also believe that increasing the training data size can improve the proposed method's performance. Additionally, as our method is modular and consists of an encoder and decoder, switching to other models such as EfficientNet \cite{tan2019efficientnet}, LSTM \cite{hochreiter1997long} GRU \cite{cho2014properties}, could be able to improve the performance of the proposed method. Ablation studies on different hyperparameters should also be conducted to determine their corresponding effect on the model. In our future work, we also plan to test more event representation methods such as Time Surface \cite{lagorce2016hots}, motion-compensated event image \cite{gallego2018unifying}, and 3D Point Set \cite{gallego2020event}.

\section{Acknowledgement}
This work was supported in part by Beijing Natural Science Foundation under Grant $4222003$, National Natural Science Foundation of China under Grant $62177001$, Research Foundation for Advanced Talents of Beijing Technology and Business University under Grant $19008022321$, and Chongqing Natural Science Foundation under Grant CSTB2022NSCQ-MSX1415.

\bibliographystyle{IEEEtran}
\bibliography{ref.bib}
\end{document}